\documentclass[sn-mathphys]{sn-jnl}

\jyear{2026}%

\theoremstyle{thmstyleone}

\theoremstyle{thmstyletwo}

\theoremstyle{thmstylethree}

\usepackage{etoolbox}
\usepackage{subcaption}
\usepackage{siunitx}
\pretocmd{\abstractname}{\newpage}{}{}

\begin{document}

\title[AI Emissions]{Greenhouse-gas emissions and the social cost of AI queries} 

\author*[1]{\fnm{Marc N.} \sur{Conte}}\email{marc.conte@fordham.edu}
\equalcont{These authors contributed equally to this work.}

\author[2]{\fnm{Djallel} \sur{Bouneffouf}}\email{Djallel.Bouneffouf@ibm.com}
\equalcont{These authors contributed equally to this work.}

\author[3,4]{\fnm{Daniel T.} \sur{Kaffine}}\email{daniel.kaffine@colorado.edu}
\equalcont{These authors contributed equally to this work.}

\author[3]{\fnm{Boyang} \sur{Yu}}\email{boyang.yu@colorado.edu}
\equalcont{These authors contributed equally to this work.}

\affil[1]{\orgdiv{Department of Economics}, \orgname{Fordham University}, \orgaddress{\street{441 E Fordham Rd.}, \city{Bronx}, \postcode{10461}, \state{New York}, \country{USA}}}

\affil[2]{\orgname{IBM Research}, \orgaddress{\street{1101 Kitchawan Road}, \city{ Yorktown Heights}, \postcode{10598}, \state{NY}, \country{USA}}}

\affil[3]{\orgdiv{Department of Economics}, \orgname{University of Colorado Boulder}, \orgaddress{\street{256 UCB}, \city{Boulder}, \postcode{80309}, \state{Colorado}, \country{USA}}}

\affil[4]{\orgdiv{Renewable and Sustainable Energy Institute}, \orgname{University of Colorado Boulder}, \orgaddress{\street{027 UCB}, \city{Boulder}, \postcode{80309}, \state{Colorado}, \country{USA}}}

\maketitle

\newpage

The rapid growth in AI usage has spurred concerns regarding the environmental impacts of electricity consumed by AI systems.  Here, we develop an integrated framework that combines experimental and theoretical estimates of per inference query electricity use with causal estimates of US regional marginal emissions rates to quantify emissions per query. We find that carbon emissions per query vary substantially across time, location, and type of model (local versus cloud). A smaller model operating in a relatively emissions-intensive electricity grid can produce greater carbon emissions than a larger model operating in a cleaner grid. We quantify the potential emission savings from altering the model used to answer the query and the time and location of the response, finding the potential to reduce emissions by roughly 85\% for identical queries within both open and closed models. We also analyze the underlying mechanisms behind the variation in emissions, connecting the milliwatt-hour AI scale of queries to the megawatt-hour scale of the electricity grid and emissions. Ultimately, AI query emissions depend on the combination of the electricity required to serve a query and the marginal carbon emissions associated with supplying that electricity.  We estimate climate damages per billion queries (a plausible estimate of 2026 daily usage in the U.S.) ranging from \$1,000-\$29,000 (2020 US dollars) for local models and \$62,000-\$649,000 (2020 US dollars) for cloud models. Extrapolating our emissions-intensity results to an assumed level of 1 quadrillion annual queries globally would project AI model inference emissions to be roughly equivalent to 10\% of 2025 annual global emissions. These results emphasize the potential scale of reductions in climate harms via routing and scheduling of AI workloads and provide a quantitative basis for carbon pricing and policy.
\\

\noindent{keywords: AI Emissions; marginal emission factors; carbon accounting; social cost of carbon; AI electricity consumption}

\newpage

AI companies have chosen to heavily subsidize the private cost of using their models, leading to substantial amounts of model interaction. From November 2024 through November 2025, there were over 100 trillion tokens worth of queries and model responses logged via a single AI inference provider, OpenRouter, offering access to a variety of large-language models (LLMs) \cite{aubakirova2025stateofai}. This suggests that there are quadrillions ($1\times10^{15}$) of tokens of AI queries and responses being conducted annually across the globe. At this scale of activity, understanding the costs and benefits of AI queries, particularly the differences between the private and social costs faced by AI users, is critical to ensure that AI is being deployed by the market in a way that is most beneficial for human well-being. While there may be many potential gaps between the private and social costs of AI use, we focus here on the social costs of climate-related damages from AI inference at the query level.

The potential for large climate-related damages from AI use is clear.  For example, after decades of limited growth, US electricity consumption has grown by 1.7\% per year since 2020, and is projected to grow further due in large part to projected AI usage \cite{eia2025electricityconsumption}. In order to address the potential social damages associated with the carbon emissions from AI-model use via carbon-aware AI policy, we must understand how consumer use of AI results in greenhouse gas emissions. A query-level accounting of the carbon footprint of AI model use aligns with the underlying behavior that gives rise to carbon damages. For example, query-level accounting can capture the climate implications of individuals' choices about which model to interact with, when and where to initiate that interaction, and how energy-intensive the interaction will be. Not all of these margins or mechanisms that determine the external costs associated with greenhouse-gas emissions are identified using other levels of accounting (such as considerations conducted at the model or data center level), as the model query is the fundamental unit of consumption in AI. 

The accumulation of trillions of query interactions has increased the sophistication and capability of AI models, which has been reflected in a dramatic shift in the type and duration of interactions between users and models. From 2024 through 2025, the number of prompt tokens per query grew four times and the output tokens per query tripled, with the number of queries related to programming increasing by roughly four times over this period  \citep{aubakirova2025stateofai}. While previous work has considered the environmental impact of AI at the query level \cite{jegham2025hungry}, it is based on a catalog of queries that do not reflect the sophistication and scope of queries being posed today. More recently, You et al. \cite{you2026quantifying} connect query-level inference energy consumption to data-center electricity demand and use a structural power-system model to simulate locational marginal carbon emissions associated with incremental AI workloads in the UK. Other studies examining the environmental impacts of data centers have relied on average emission rates from observed data, either at broad geographic scales \cite{muller2026measuring, bistline2026aiemissions, nguyen2024towards} or by assigning balancing-authority average emission rates to individual data centers based on their location \cite{guidi2026assessing}. 

While these studies of the environmental effects of data centers are informative, the  fundamental unit of AI consumption is the query. Economic frameworks used to develop policy guidance, such as the imposition of a levy on behaviors that generate external costs, consider  tradeoffs at the margin of choice, which is not available at the resolution of a data center.  Furthermore, inference does not necessarily have to occur in a data center, and the last year has seen a dramatic increase in the usage of local models that can be run either remotely or on the user's individual device \cite{aubakirova2025stateofai}. Moreover, causal estimates of marginal emission rates have been shown to be important in other related contexts, such as electric vehicles \cite{holland2016there,holland2022marginal},  renewable energy \cite{Novan:15,CFM:2018,fell2026spatio} and AI training \cite{dodge2022measuring}. Taken together, the computational requirements of individual queries vary substantially across models and tasks, while the climate damages associated with the electricity used to serve them depend on marginal rather than average emissions. Thus, there is a pressing need for empirically grounded, query-level estimates of climate damages that account for both variation in computational requirements across models and tasks and variation in marginal emissions across electricity systems. Furthermore, a comprehensive accounting of the mechanisms behind the variability in climate damages from AI queries is necessary to identify potential paths forward to mitigate these damages while ensuring access to the benefits from AI.

In this paper, we provide a comprehensive accounting of the determinants of the greenhouse-gas emissions associated with a query sent to an AI model. We combine the results of a primary-data collection experiment and theoretical analysis measuring energy usage from a set of local and cloud AI models with an econometric analysis of the carbon emissions associated with a marginal unit of electricity production at each of the balancing authorities in the United States. To measure the energy usage, we directly measure energy usage per query from a suite of local models and derive energy usage per query from frontier cloud models.  For the econometric analysis, we use hourly energy consumption and emissions data from each balancing authority in the U.S. to causally estimate balancing-authority specific marginal emission rates by hour.  Because the fuel mix of the electricity is a critical determinant of emissions, we also estimate electricity generation responses to electricity demand.  Merging these analyses allows us to examine how carbon emissions vary based on the query type, the model used, the output generated, and the time and location where the query is handled.  We also analyze the underlying mechanisms behind these results, connecting the milliwatt-hour AI scale of queries to the megawatt-hour scale of the electricity grid, generation, and emissions. 

We find that the greenhouse-gas emissions associated with a query posed to an AI model can vary by several orders of magnitude. Within local models, model size, as measured by the number of active parameters, is the largest source of variation, with the least-energy intensive model generating 73\% fewer emissions than the most-energy intensive model. For the cloud models (on GPU arrays housed in AI factories), there is less variation across models, and the biggest source of variation in emissions for the average query comes from the location of the AI factory (85\% reduction), based on the fuel mix within the particular balancing authority at a given hour of the day. Finally, while our methods of estimating energy-use vary across local and cloud models, we note that answering a query via the most energy-efficient local model as opposed to the most energy-hungry cloud model can reduce the resulting emissions by 98\%. These results emphasize the potential scale of reductions in climate damages via routing and scheduling of AI workloads and provide a quantitative basis for carbon pricing and policy.

\section*{Results}\label{results}
\subsection*{AI Emissions and potential for emission reductions}\label{sec3a}

The carbon emissions associated with AI queries vary substantially across both models and balancing authorities (Fig. 1). Electricity demand is evaluated for 14 standardized benchmark queries spanning a range of text creation tasks that vary in reasoning and required digestion of different lengths of input content, with query complexity and output requirements ranging from short- to long-form responses. The carbon emissions for these queries are then calculated for each of 43 balancing authorities in the US with sufficient data, based on the marginal emissions rate (see Methods).  For the local models,  emissions per query generally increase with model size, but there is also substantial variation for a given model across balancing authorities (Fig. 1a). This geographic variation can be sufficiently large to outweigh differences across model sizes.  For example, consider three large balancing authorities that  vary widely in their fuel mix.  Bonneville Power Authority relies heavily on hydropower, so a query answered by the largest local model in BPAT generates much less emissions than the same query answered by the smallest local model in the more gas- and coal-dominated Pennsylvania-New Jersey-Maryland Interconnection (PJM). The Electric Reliability Council of Texas (ERCOT) is cleaner than PJM and dirtier than BPAT, such that a large model in ERCOT is dirtier than a small model in PJM, but a small model in ERCOT is roughly equivalent to a large model in BPAT. A similar degree of geographic heterogeneity is evident for the cloud models (Fig. 1b). Thus, model size is an important determinant of query-level emissions, but it does not uniquely determine them. Clearly, the electricity system supplying the computation can substantially alter the emissions associated with a given amount of AI activity.

\begin{figure}
    \centering
    \includegraphics[scale=.18]{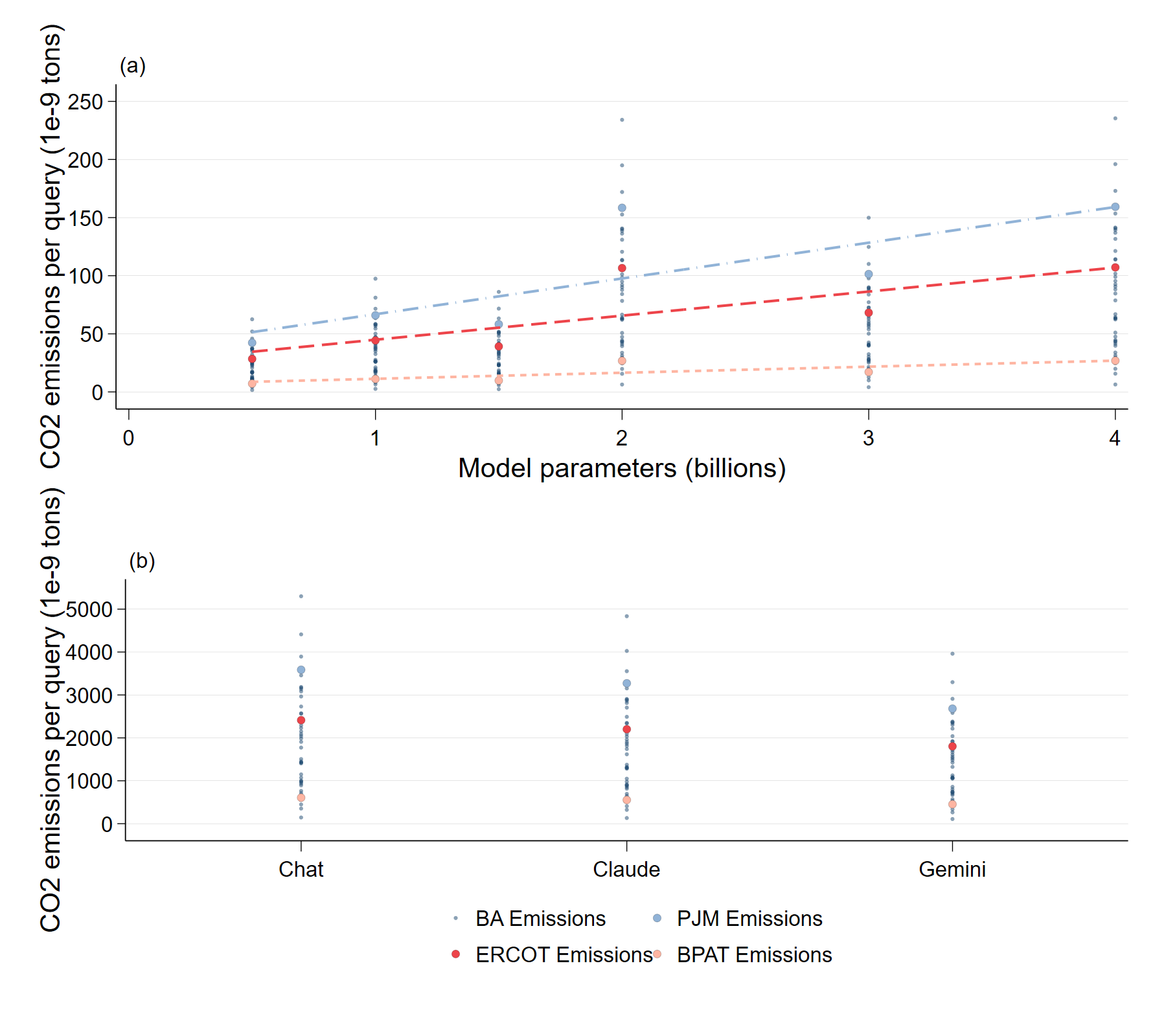}
    \caption{Carbon emissions per query across models and balancing authorities. Each point on the scatter plots represents emissions from a balancing authority for a query of that model. Panel a) presents local models, ordered by parameter size (in billions).  Panel b) presents cloud models in no particular order.}
    \label{fig:2}
\end{figure}

The magnitude of the variation in carbon emissions per query implies considerable potential for reducing query-level emissions. To quantify this potential, we construct a sequence of counterfactual scenarios that progressively alter the time, model, and location used to serve a query (Fig. 2). For local models (Fig 2a), beginning with a query served in PJM at 8 p.m. using the most emissions-intensive model considered, shifting computation to the least emissions-intensive hour reduces emissions by 14\%. Switching to the least emissions-intensive local model produces a considerably larger reduction of 73\%, while shifting the computation to BPAT lowers emissions by an additional 85\%, for a total reduction of 96.6\% relative to the initial scenario. Most of the potential reduction therefore arises from model and location choices, while shifting the timing of computation within the same balancing authority produces only modest reductions. The relative importance of these margins differs for cloud models (Fig. 2b). Shifting the query to the least emissions-intensive hour in PJM reduces emissions by 14\%, while choosing the least-emissions intensive cloud models reduces emissions by 25\%. Serving the query in BPAT produces a much larger reduction of 85\%, for a total reduction of 90.4\% relative to the initial scenario for cloud models. Geographic differences in marginal electricity emissions therefore account for a particularly large share of the potential emissions reductions for cloud models in our scenarios. Finally, allowing the query to be served by a local rather than cloud model produces a massive reduction of 98\%, though there may be important differences in the ``quality" of the responses that limit the ability to directly compare between the two model types (see Discussion below).

\begin{figure}
    \centering
    \includegraphics[scale=.70]{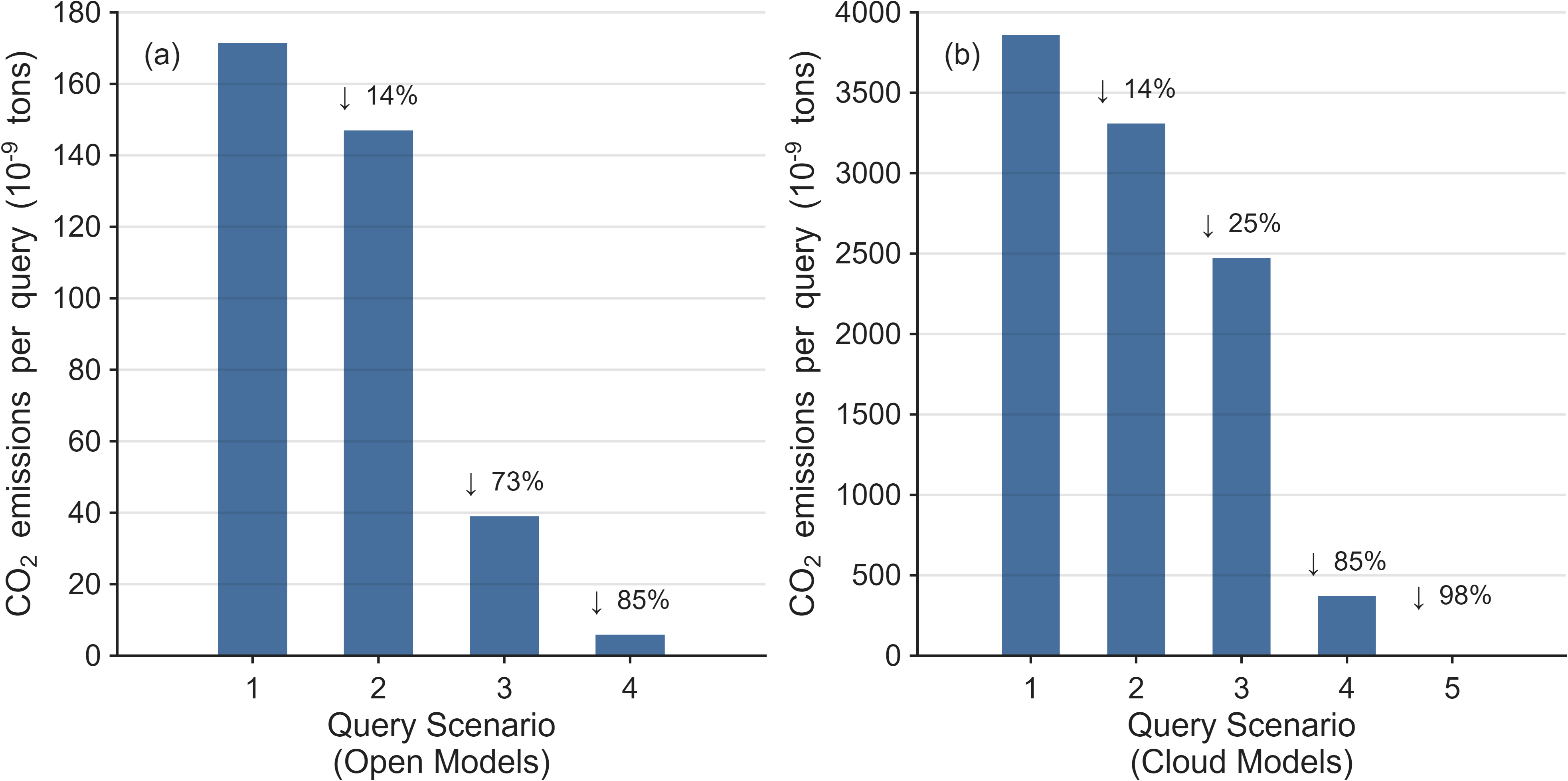}
    \caption{Carbon emissions for different query scenarios.  Scenario 1 represents the carbon emissions from a query in PJM run at 8pm on the most emissions-intensive models by type (4 billion parameters for local models, ChatGPT for cloud models). Scenario 2 answers the query at the least emissions-intensive time of day.  Scenario 3 answers the query in the same way as Scenario 2 but with the least-emissions intensive model by type. Scenario 4 answers the query in the same way as Scenario 3 but in BPAT. Finally, for the cloud models, Scenario 5 answers the query in the same way as Scenario 3 but on a local model.}
    \label{fig:3}
\end{figure}

These carbon emissions can be converted into climate damages via the social cost of carbon, which we take as \$185 per metric ton of CO$_2$ in 2020 US dollars \cite{rennert2022comprehensive}. As of July 2025, ChatGPT alone was receiving more than 330 million queries per day in the U.S. \cite{openai2025economicanalysis}, suggesting total U.S queries could plausibly be on the order of a billion queries per day.  For local models, climate damages per billion queries range from \$984 in Scenario 4 to \$28,808 in Scenario 1. Similarly, for cloud models, climate damages per billion queries ranges from \$62,319 in Scenario 4 to \$648,580 in Scenario 1.  To put that into perspective, for a frequent individual user who generates 10,000 prompts per year (roughly 30 a day), their climate damages can range from a tiny \$0.01 per year for small local models in clean locations to \$6.48 per year for cloud models in dirty locations.  At the individual level, these climate damages may appear modest, but the nearly three-orders-of-magnitude difference can become substantial in aggregate. For example, if every U.S. resident generated 10,000 prompts per year, aggregate climate damages would range from \$3.4 million annually compared to \$2.3 billion annually. 

\subsection*{Mechanisms}\label{sec3}
The large variation in query-level emissions shown in Fig. 1 reflects heterogeneity in the electricity required to serve a query and the marginal carbon emissions associated with supplying that electricity. We first examine the computational mechanisms underlying differences in energy use (Fig. 3).

\begin{figure}
\centering
\begin{subfigure}{.5\textwidth}
  \centering
  \includegraphics[width=\linewidth]{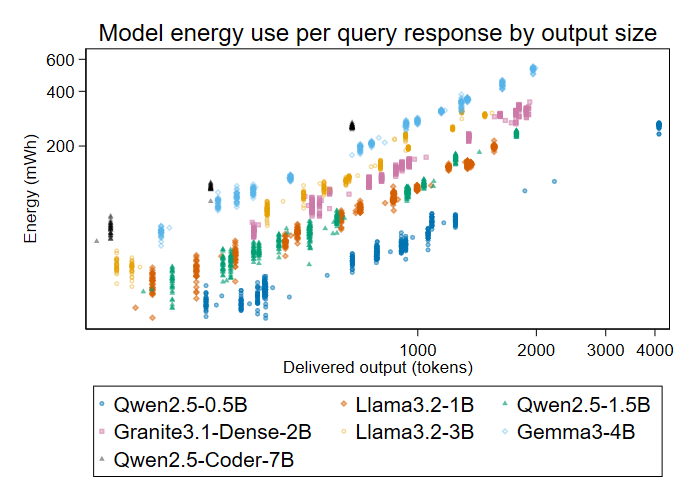}
  \includegraphics[width=\linewidth]{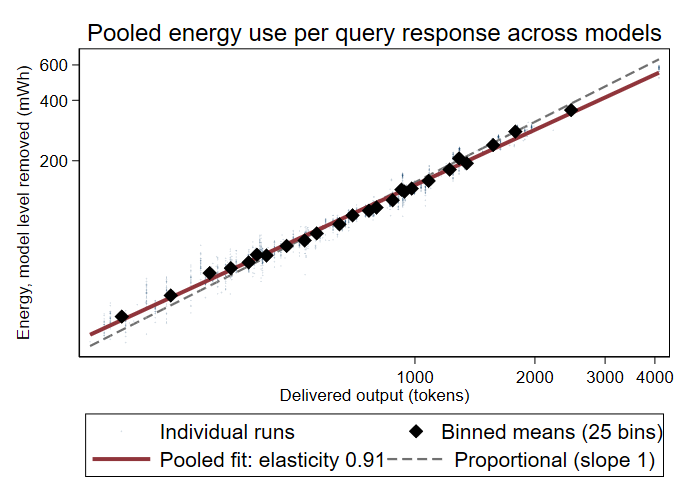}
  \caption{Results for Local Models}
  \label{fig:sub1}
\end{subfigure}%
\begin{subfigure}{.5\textwidth}
  \centering
  \includegraphics[width=\linewidth]{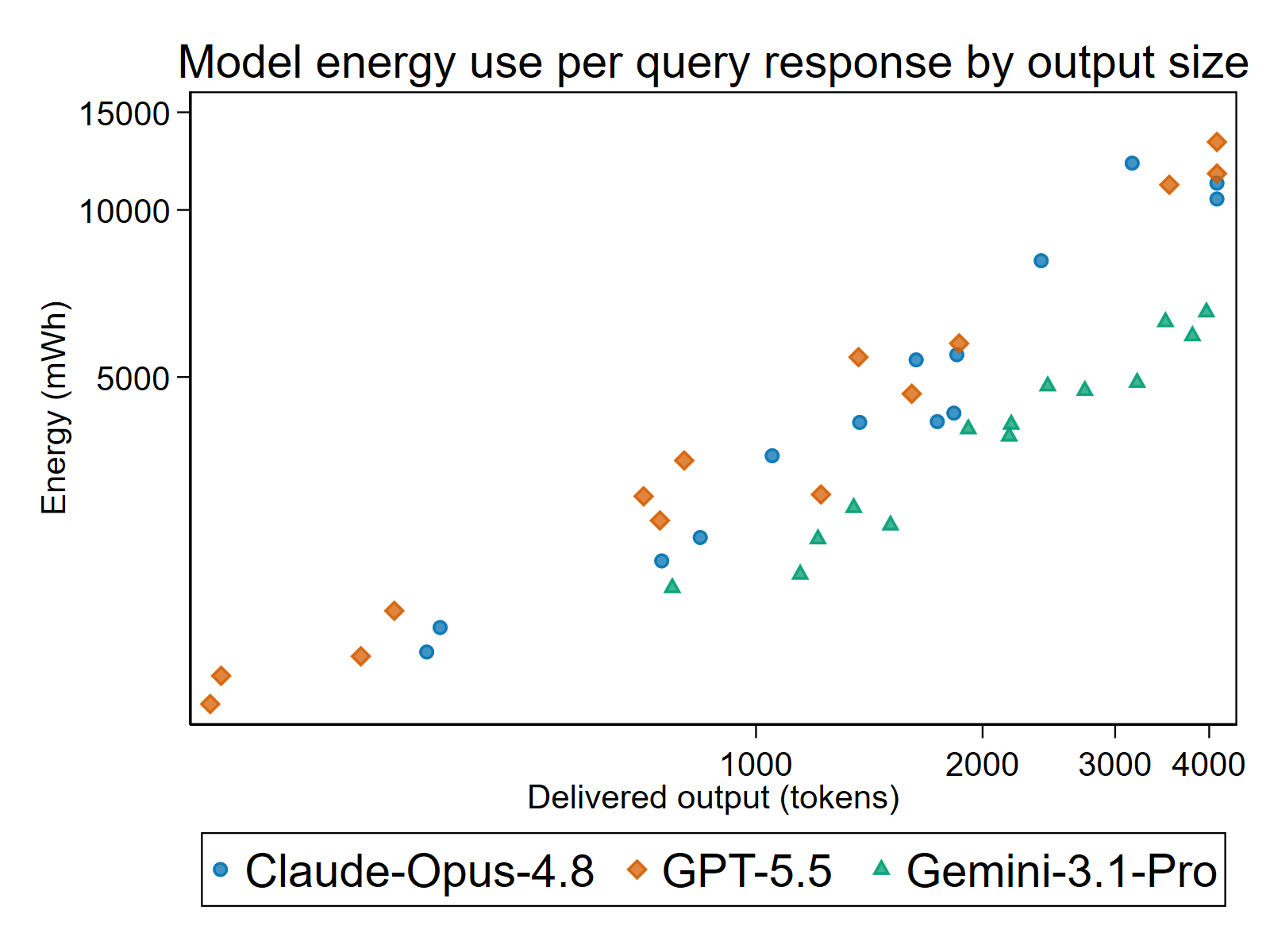}
    \includegraphics[width=\linewidth]{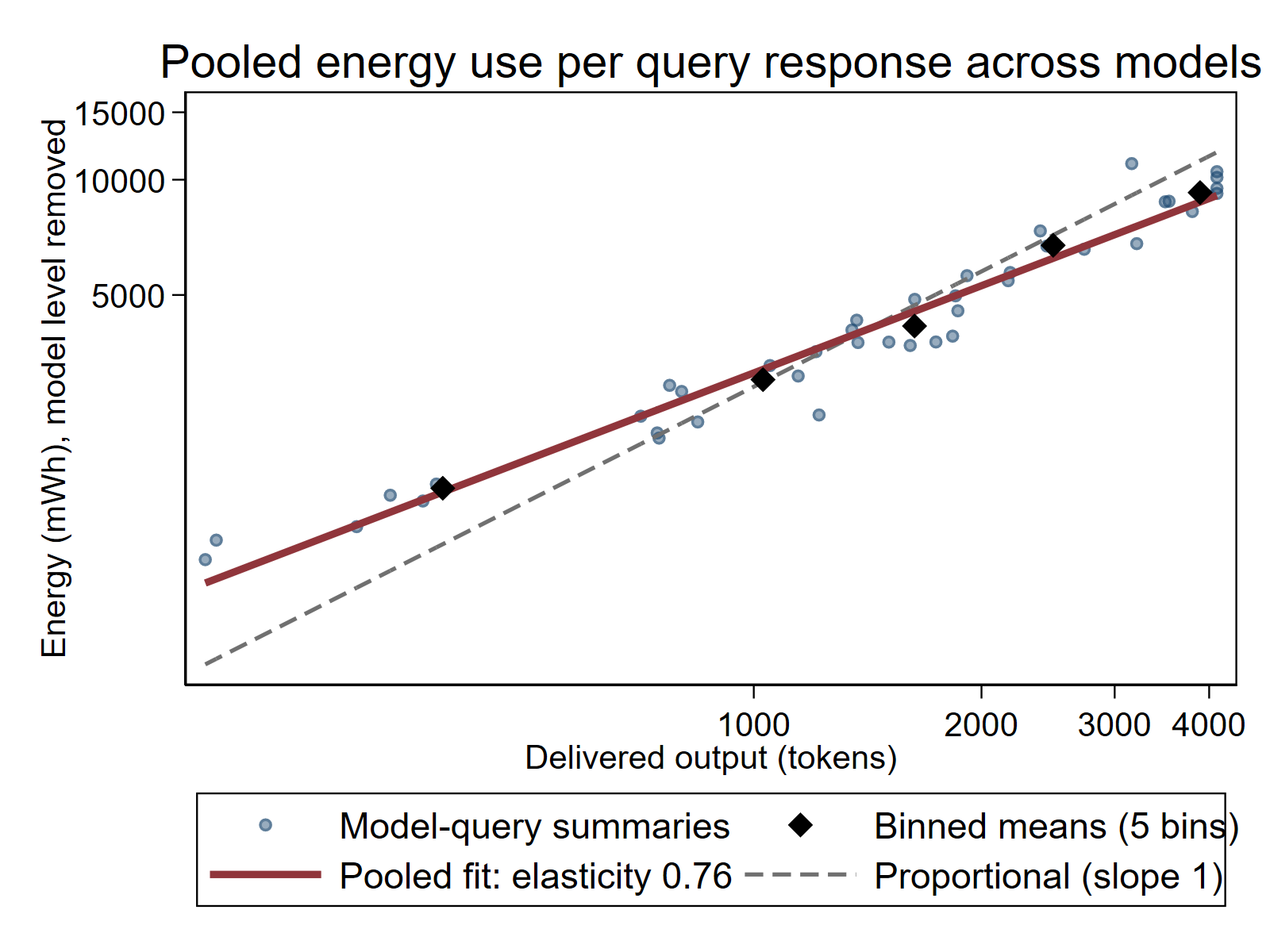}
  \caption{Results for Cloud Models}
  \label{fig:sub2}
\end{subfigure}
\caption{Column a) presents results for local models. The top panel (a1) displays the amount of energy used per query as measured in milliwatt-hours (mWh) by each of the seven tested models across a range of delivered output sizes, as measured in tokens. The lower panel (a2) aggregates query responses and absorbs model-specific energy intensities (model fixed effects $\alpha_{m(i)}$ in the regression $\ln E_i \;=\; \alpha_{m(i)} \;+\; \beta \,\ln Y_i \;+\; \varepsilon_i$) to estimate how much energy each query would have required if handled by the average model. The solid line is the pooled fit, $\exp(\bar{\alpha} + \hat{\beta}\ln Y)$, and the dashed line is a reference with a slope of one (implying energy usage is proportional to output). The estimated regression imposes a common slope, so the bottom figure in the column assumes the same elasticity for every model, while the top figure in the column shows the unadjusted data for comparison. Column b) presents the same results for the cloud models in our study in panels b1 and b2, respectively.}  
\label{fig:three}
\end{figure}

The number of output tokens generated in response to the benchmark queries varies substantially across queries and models for both local (Fig. 3a1) and cloud models (Fig. 3b1). This variation partly reflects the deliberately broad range of tasks in our benchmark set, which includes short-form writing and presentation tasks as well as substantially longer synthesis exercises (see SI 1 for query descriptions and details). The output length matters directly for electricity consumption, as within models, total energy use increases approximately linearly with the number of tokens generated (Fig. 3a1 and 3b1). Thus, differences in the quantity of output generated can produce substantial variation in the energy required to answer different queries, even when the underlying model is unchanged.

Energy use also varies systematically across models. For local models, energy consumption per output token generally increases with model size, as measured by active parameters (Fig. 3a1), providing a second channel through which larger models tend to consume more electricity per query. While our methods to estimate energy demand differ across local and cloud models, we can see from a comparison of figures 3a1 and 3b1 that the delivered output by these two categories of models in response to the same set of queries differed somewhat substantially. For example, 72.3\% of the query responses are below 1000 tokens for  local models, while only 42.9\% of responses are in that output range for the three cloud models. However, this average across the three cloud models masks substantial heterogeneity, with 77.8\% of GPT-5.5 query responses, 50\% of Claude-Opus-4.8 query responses, and only 9.1\% of Gemini-3.1-Pro query responses including fewer than 1000 tokens. 

The model-specific figures (3a1 and 3b1) show how each model’s energy use varies with the length of its responses; however, any comparisons across differences in model intensity are confounded by differences in model verbosity. Figure 3 panels a2 and b2 present results of pooled specifications that separate these two characteristics. Here, model fixed effects capture each model’s energy level at a common output length, and the single common slope estimates how energy use increases with delivered output.  
We estimate an elasticity of 0.91 for the local models and 0.76 for the cloud models, indicating that energy rises less than proportionally with output and that the difference appears larger for the cloud models, subject to the assumption that slopes are common across models within each group. 

These relationships combine to generate the substantial differences in total energy use observed across models and queries. Models differ in the electricity required to generate a given amount of output, while benchmark queries differ considerably in how much output they generate. As a result, query-level electricity consumption depends on both the model choice and generated output.

Turning to the megawatt-hour scale of the grid, electricity system conditions are the second source of heterogeneity in query-level emissions (Fig. 4). Marginal CO$_2$ emission rates vary considerably across US balancing authorities, with relatively low rates across portions of the western United States and substantially higher rates in many central and eastern grids (Fig. 4, top panel). The same quantity of additional electricity demand can therefore produce substantially different emissions depending on where it occurs.  Importantly, these estimates represent the marginal emissions rate in a balancing authority (see Methods), which can be substantially different from the average emissions rate.

\begin{figure}
    \centering
    \includegraphics[width=0.95\linewidth]{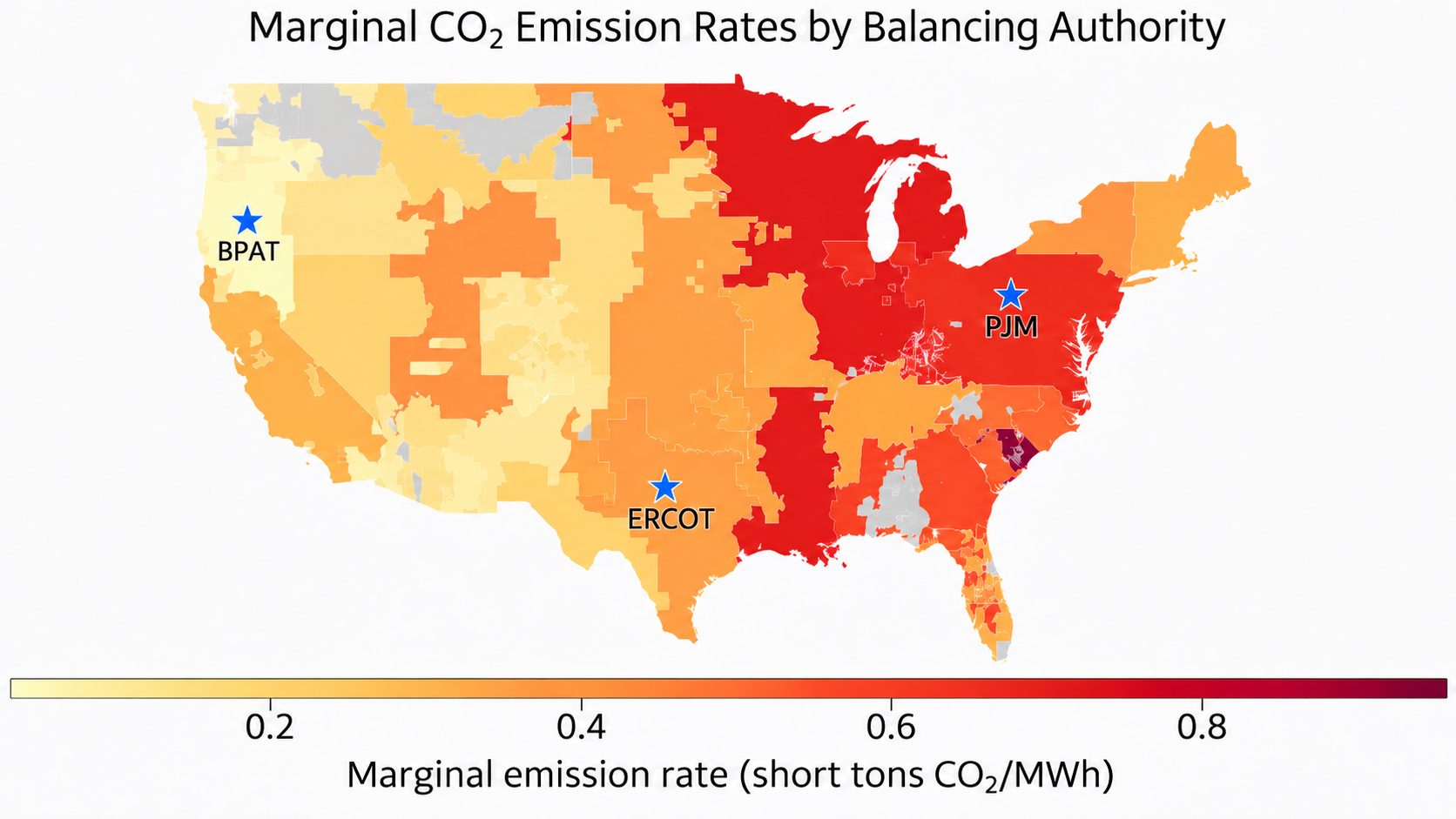}
   \vspace{0.25cm}

    \includegraphics[scale=.2]{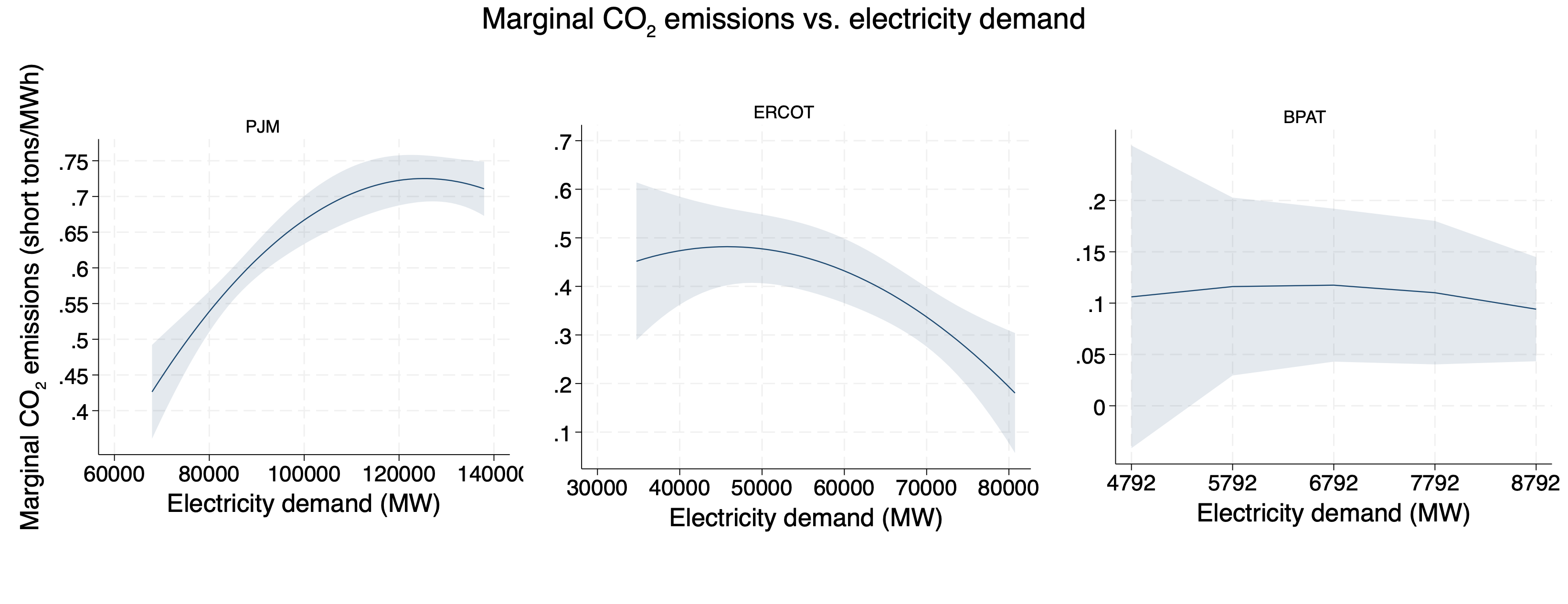}
    
    \caption{Top panel: Colors show load-weighted  marginal CO$_2$ emission rates by balancing authority. Gray areas indicate balancing authorities with insufficient data to generate estimates. Rates are measured in short tons of CO$_2$ per MWh. Bottom panel: Marginal CO$_2$ emission rates as a function of electricity demand for PJM, ERCOT, and BPAT. Marginal emission rates are estimated from cubic regressions of hourly CO$_2$ emissions on electricity load, including year$\times$month, month$\times$hour, and year$\times$hour fixed effects. The solid lines show the estimated marginal effect of a 1 MWh increase in load at different demand levels, and shaded areas denote 95\% confidence intervals. Standard errors are clustered by year-month.}
    \label{fig:5(a)}
\end{figure}

This cross-sectional variation across balancing authorities reflects differences in the underlying generation resources used to meet incremental electricity demand. The relationship between total system demand and marginal emissions differs markedly across PJM, ERCOT, and BPAT (Fig. 4, bottom panel - note the vertical scale). In PJM, marginal emissions rise with demand over much of the observed load range before flattening at high loads. In ERCOT, marginal emissions instead decline as system demand increases, while marginal emissions in BPAT are both substantially lower and comparatively flat. 

These patterns are consistent with the distinct mix of generation that responds to incremental demand in each electricity system over the course of the day (Fig. 5, top panel - note the vertical scale), which then drives the changes in emission rates. PJM exhibits relatively high marginal emissions overnight and during evening hours and lower marginal emissions during the morning and mid-day. By contrast, ERCOT has lower emission rates overnight, and highest emission rates in the pre-dawn early morning hours. BPAT remains comparatively low-emitting across hours, although its marginal emissions also vary over the day.

These marginal emission rates can be explained by fuel-specific generation responses (Fig. 5, bottom panel). In PJM, higher emission coal ($\approx 1.0$ tons of CO$_2$ per megawatt-hour) provides a relatively larger share of the marginal generation response overnight, while relatively cleaner natural gas ($\approx 0.4-0.6$ tons of CO$_2$ per megawatt-hour) becomes more important during morning hours. ERCOT shows a different pattern, with wind frequently contributing to incremental generation overnight and natural gas and clean solar generation responding during daylight hours. In BPAT, clean hydropower accounts for much of the response to incremental demand across all hours of the day. As shown by these examples, the emission rate associated with the marginal generation unit that responds to additional electricity demand is the key source of variation in the map in Fig 4.  In balancing authorities where coal and natural gas are typically the marginal unit, marginal emissions rates are higher, while in balancing authorities where wind, solar or hydro are frequently the marginal unit, marginal emissions rates can be much lower. 

\begin{figure}
    \centering
    \includegraphics[width=\textwidth]{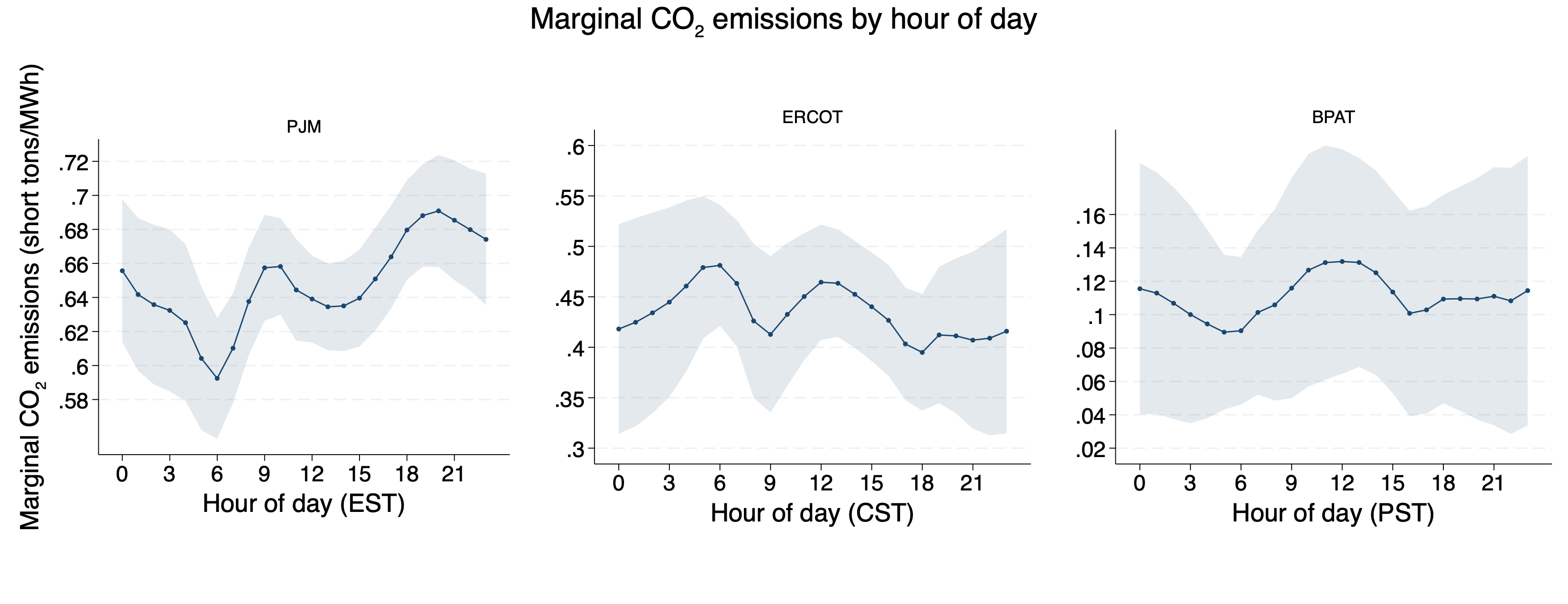}

\vspace{0.25cm}

    \includegraphics[scale=.2]{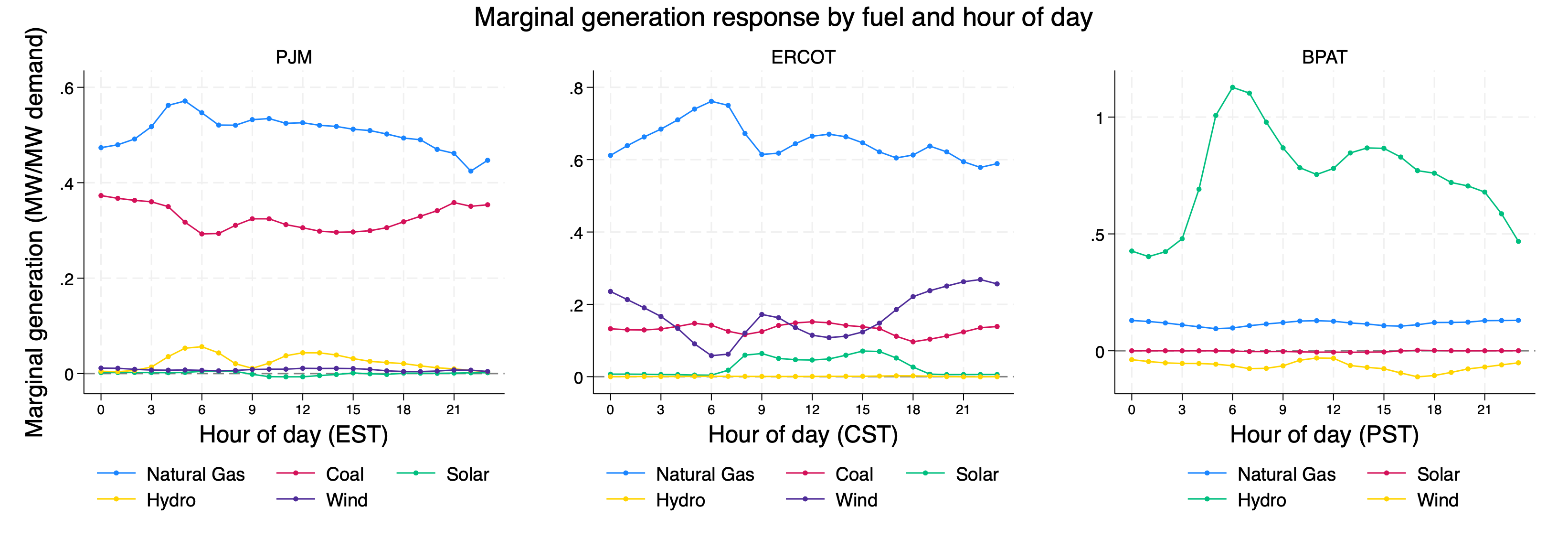}
    
    \caption{Top panel: Points represent estimated hour-specific marginal CO$_2$ emission rates for three balancing authorities: PJM, ERCOT, and BPAT. Shaded areas denote 95\% confidence intervals. Regressions include year$\times$month, month$\times$hour, and year$\times$hour fixed effects, with standard errors clustered by year-month. Hours are reported in local standard time. Bottom panel: Lines show estimated fuel-specific marginal generation responses to a 1 MW increase in system demand for PJM, ERCOT, and BPAT. Regressions include year$\times$month, month$\times$hour, and year$\times$hour fixed effects, with standard errors clustered by year-month. Hours are reported in local standard time. 
    }
    \label{fig:5(c)}
\end{figure}

Together, the computational and electricity-system mechanisms described above explain the broad dispersion in query-level emissions shown in Fig. 1. Model choice and generated output determine the electricity required to serve a query, while location and time of day determine the marginal emissions associated with that electricity consumption. Variation along either dimension can significantly change query-level emissions, and the interaction between them accounts for the large differences in emissions per query across the model-location combinations, per Fig. 2.

\section*{Discussion}\label{sec4}

The state of technology related to AI and its environmental impacts is changing rapidly (e.g., preferred cooling technologies, model design, model use). Our analysis advances our understanding in this area by estimating energy demand at the query level for a range of query types that approximate current usage, and then mapping those energy demands to causal emission estimates across U.S. balancing authorities. Our query-level approach offers an advantage relative to data center-level accounting that has been followed in existing work, as the query-level approach allows for development of a market intervention that could better address the negative environmental impacts of AI model use, while also acknowledging the social benefits associated with AI.

Our results highlight that marginal emissions can vary significantly based on the query posed, model used, and the time and location of query processing. Among local models, we find that emissions are generally increasing in model size, as measured by active parameters, although the geographic variation in balancing-authority emissions intensity based on fuel-mix is sufficiently large that a large model run in a clean location can generate fewer emissions than a small model run in a relatively dirty location. The variation in marginal carbon emissions across balancing authorities has an impact of a similar magnitude for cloud-based models.  While the $CO_2$ emissions resulting from a single model query are small, on the order of 0.1 grams (local) to 1 gram (cloud), the sheer magnitude of queries annually means that understanding these effects is material. A back-of-the-envelope calculation scaling our cloud model results at current estimated emissions per query to global AI usage would imply roughly 1 quadrillion annual queries would generate the equivalent of 10\% of global 2025 CO$_2$ emissions ($\approx$ 5 gigatons). Handling these queries on local models could reduce emissions by 25-fold. Our estimate here is conservative for two reasons. First, we only measure the electricity required for model inference, which should reflect less than half of full model energy demand for cloud models that includes training and electricity for cooling as well \citep{jegham2025hungry}. Second, we assume that all of the electricity used to respond to AI model queries is provided by the electrical grid, which is not necessarily the case for cloud models.  A recent report has shown that behind-the-meter energy sources relied on by AI companies tend to use the cheapest/least-efficient forms of gas plants (simple-cycle or reciprocating internal combustion engines, which both have higher heat rates than combined-cycle gas plants \citep{rhodium_2026}). Relying on behind-the-meter sources exclusively would increase our projections of annual CO$_2$ emissions from 1 quadrillion model queries  to 13-19\% of global 2025 CO$_2$ emissions.  

A core lesson of economic analysis is that efficient policy interventions address the root cause of the external costs - that is, the difference between the private and social costs of behavior. For AI users to internalize these external costs, they must be reflected at the relevant margin of consumption, which for AI is the model query.  As such, two conditions for efficient AI use are required to hold: 1) electricity costs must be embodied in the price paid per query by AI consumers, and 2) carbon pricing must be embodied in those electricity costs at levels consistent with the social costs of carbon \cite{rennert2022comprehensive}. In principle, this would lead AI consumers to internalize the carbon damage associated with each query.  In practice however, these conditions do not hold in most of the world  and are unlikely to do so in the near future.  For example, in the U.S., the external costs of electricity consumption are as large (or larger) than the private costs for most of the country (Fig 6 - see Methods).  In the largest balancing authority, PJM, external costs are nearly double private costs. 

\begin{figure}
    \centering
    \includegraphics[width=0.95\linewidth]{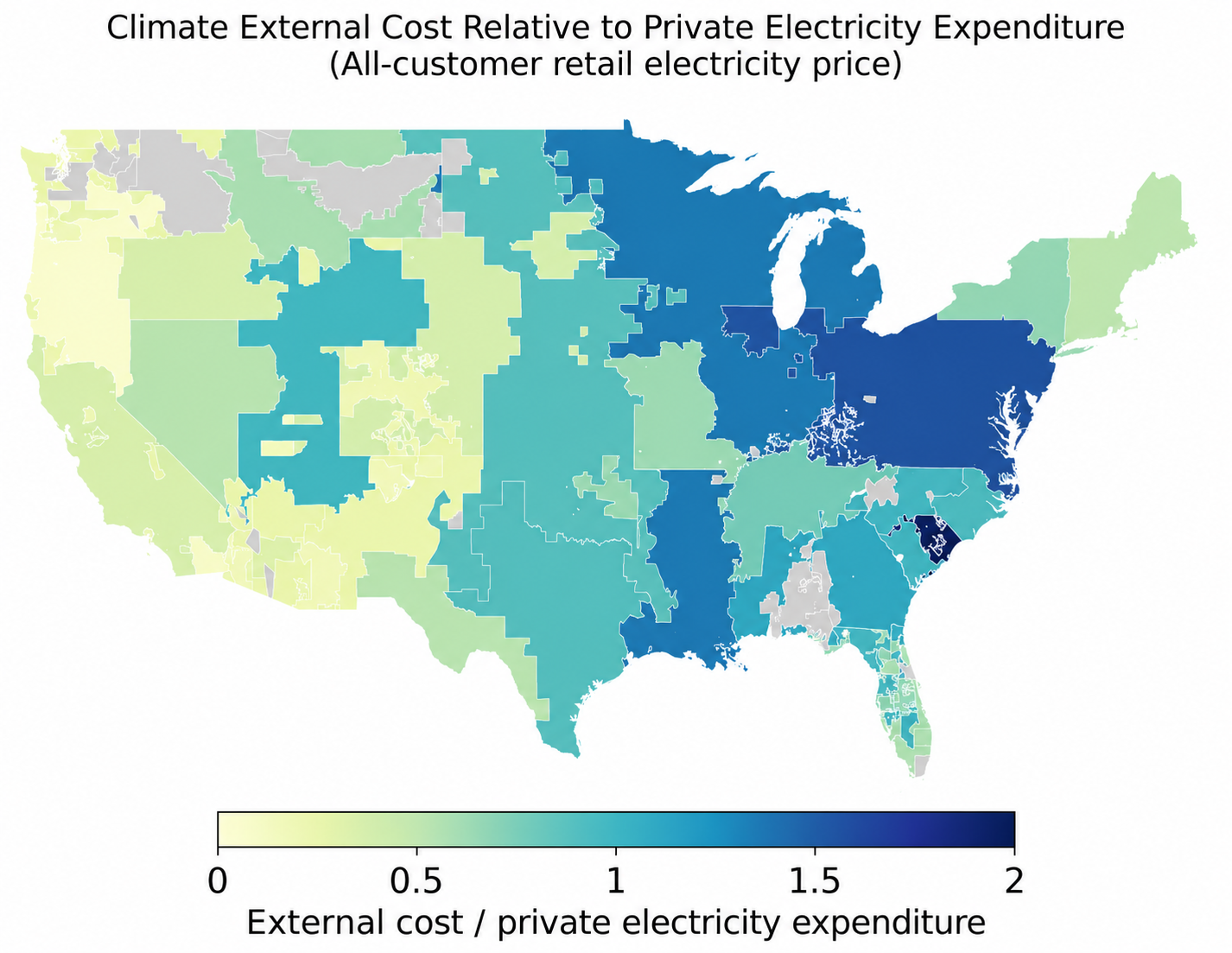}
    \caption{Climate external cost relative to private electricity expenditure across U.S. balancing authorities, using 2024 average retail electricity prices across all customer classes.}
    \label{fig:8(a)}
\end{figure}

In the absence of comprehensive carbon pricing, how might we implement the emission reductions from Figure 2? The results suggest several categories of actions that could reduce the carbon footprint of AI (See SI 3). The first action is to choose less compute for each query, as our results show that smaller models should be used when they achieve comparable performance, and model orchestration can match the model to the query. The second action is to reduce the number of tokens processed and generated, either via user-controlled query size restraints or model-side restrictions to shorten output. However, our results offer some caution about the extent to which models will adhere to directives regarding output length. The third action is to make each computation use less energy via model efficiencies, such as quantization, pruning, and distillation, or optimized inference via caching, batching, and efficient attention mechanisms.  Of course, each technique must be checked for accuracy loss and quality control.  Finally, the variation in marginal emission rates across space and time implies gains from orchestration across when and where queries are processed. Typical AI consumers will be neither incentivized nor have sufficient knowledge to orchestrate their own queries, and so this effort would likely require AI firms to conduct the orchestration. Our suggestions regarding model orchestration to coordinate the timing and location of query response to minimize the carbon footprint echo the points in \citet{dodge2022measuring}. Furthermore, our analysis of an array of models and queries expands our recommendations to include a number of additional actions related to the demand and supply-side of AI tools that we find to be potentially more impactful than the timing- and location-related adjustments. 

We acknowledge several limitations of our analysis.  First, it is important to emphasize that our estimates to this point have been based solely on the energy required for the AI models to respond to each query. A comprehensive query-level accounting of energy use would need to also include the energy needed to cool the GPUs running the various AI models as well as the energy used to train these models so that they are able to provide quality responses to queries \cite{jegham2025hungry}.  Second, we estimate marginal emission responses to queries, which takes the current grid as given, and reflects the expected damages of an individual AI query by a single user today.  Of course, widespread adoption of AI or concentration of data centers in certain locations will necessitate substantial changes in generation supply, with implications for emissions \cite{gagnon2022short}. For example, a large build-out of combined cycle natural gas generators with emission intensities of 0.4 tons/MWh could look cleaner in PJM, but dirtier in BPAT.  Finally, throughout the analysis, we have generally kept the results for the cloud models separate from those for the local models.  The key rationale is that while cloud models were substantially more emissions-intensive, it is unclear if the quality of query responses is comparable between local and cloud models. Future research can consider whether the potential improvements in inference quality are worth the energy and emissions costs.

\section*{Methods}

\subsection*{Conceptual framework}
To connect AI queries to carbon emissions, we start with the link between AI queries and energy demand. When users type a query into their computer, the text is broken into small pieces called tokens (\textit{tokenized}), which can be words, parts of words, or symbols, before it can be processed.  These tokens are converted into numerical representations called embeddings and passed through a neural network. In a transformer-based model (e.g. an AI model like ChatGPT), the neural network performs billions or trillions of mathematical operations, primarily matrix multiplications, to predict the next token. The model generates the response one token at a time, and these tokens are converted back into text and displayed on the user's computer. Every mathematical operation performed by the model requires electricity because it runs on Graphical Processing Units (GPUs) or other specialized AI accelerators inside a data center. Electricity is used not only by the GPUs performing the computation, but also by memory, networking, storage, and cooling systems. Ultimately, the amount of electricity used per query depends on factors such as the model's size, the number of input and output tokens, the hardware being used, and how efficiently the GPU or data center operates. 

To go from the milliwatt-hour scale to the megawatt-hour scale, the demand for electricity to meet an AI query must be met by supply from the electric grid.  We consider balancing authorities as our unit of observation (instead of utilities or states), as they are the formal entity tasked with ensuring the balance of electricity supply and demand within their spatial footprint. Balancing authorities in the United States vary substantially in size. There are several large, multi-state regional transmission operators (RTOs) such as the Midcontinent Independent System Operator (MISO),  Pennsylvania-New Jersey-Maryland Interconnection (PJM), or the Southwest Power Pool (SPP).  Other balancing authorities serve a large percentage of a single state such as the Electric Reliability Council of Texas (ERCOT), Public Service Company of Colorado (PSCO), or Nevada Power (NEVP). Finally, some balancing authorities serve only a small portion of a state, such as the Los Angeles Department of Water and Power (LADWP) and the many small balancing authorities that serve parts of Florida.

Across the US, electricity is dispatched through different market and regulatory structures, but with a similar underlying economic logic. Generators are ordered by cost (or bid) to form a supply curve (the \textit{merit order}), and demand is met by moving along that supply curve. The key feature is that for any incremental unit of electricity demand (load), there exists some generator - the \textit{marginal generator} - in the merit order that will increase its output. For balancing authorities participating in wholesale markets, uniform-price auctions are held where generators submit price-quantity bids to form the supply curve. At any point in time, the intersection of demand with this supply curve determines the market-clearing price that all dispatched generators receive, and all generators with bids at or below that price are dispatched.  Balancing authorities with vertically integrated utilities engage in a similar exercise, but use internal data on costs to construct the supply curve and determine least-cost dispatch to meet demand.   

The emissions from the marginal generator will ultimately determine the marginal emissions generated by serving an AI query.  Fossil fuel CO$_2$ emissions from combustion range from approximately 1.0 metric tons/MWh for coal, to 0.8 metric tons/MWh for diesel and fuel oil, to 0.6 metric tons/MWh for simple-cycle natural gas, and down to 0.4 metric tons/MWh or less for combined-cycle natural gas \cite{epa2020egrid}. Crucially, the marginal emission rate can be quite different than the average emissions (often referred to as the emissions intensity) \cite{Kaff:2013,holland2022marginal,bajo2025decarbonization}.  For example, a balancing authority with many clean but relatively inflexible nuclear plants may have a low emissions intensity (average emissions) but a higher marginal emission rate if flexible natural gas responds to incremental changes in electricity demand.

\subsection*{AI model energy use}
To determine the energy demand per query, we directly meter energy usage for local models and derive energy usage for cloud models. For the local models, six dense instruction-tuned models spanning \SI{0.5}{B}--\SI{4}{B} parameters are served by Ollama on a single NVIDIA RTX 4000 Ada Generation Laptop GPU, with one prompt at a time, greedy decoding ($T = 0$), and \verb|num_predict| = \num{4096}. Each prompt cell is repeated 30 times, giving $6 \times 14 \times 30 = \num{2520}$ scheduled runs. Note that Qwen2.5-Coder-7B with 7 billion parameters was also included in the study, but it generated an incomplete set of responses and results are only reported in Figure 3 for the partial set of responses.

Board power ($P(t)$) is directly sampled in a background thread at \SI{0.3}{s} intervals via \texttt{nvidia-smi}.  Power usage is compared against a 5-second no-load baseline that is taken once at start-up ($P_{\mathrm{idle}} = \SI{4.13}{W}$). For a request of duration $T_q$ with $K_q$ power samples,
\begin{equation}
E_q \;=\; \frac{\overline{P_q}\,T_q}{3600}, \qquad
\overline{P_q} = \frac{1}{K_q}\sum_{k=1}^{K_q} P(t_k),
\qquad\quad
E_q^{\mathrm{marg}} \;=\; \frac{\bigl(\overline{P_q}-P_{\mathrm{idle}}\bigr)^{+} T_q}{3600},
\label{eq:local}
\end{equation}
in watt-hours (Wh). We use total GPU board energy per query $E_q$ throughout the paper.  The marginal energy per query $E_q^{\mathrm{marg}}$  implies that inference accounts for \SI{92.3}{\percent} of energy usage on average, so our results are not sensitive to the inclusion or exclusion of idle energy. Observed board power ranges from the \SI{4.13}{W} idle floor to a \SI{110.8}{W} peak. Finally, token counts come from Ollama's \verb|eval_count| / \verb|prompt_eval_count| fields. 

We drop failed calls and any run whose duration exceeds $5\times$ its own cell's \emph{median} duration, as we interpret those long runs as a stall. We use the median as the reference  because it is unaffected by the outlier being tested and the $5\times$ threshold is much larger than the underlying within-cell variation of the surviving runs (none of the coefficients of variation on duration exceed \SI{30}{\percent} and the 99th percentile is \SI{29.1}{\percent}). Of \num{2520} attempted local runs, 6 failed and 2 were stalls (\SI{503}{s} and \SI{938}{s} compared to cell medians near \SI{10}{s}), leaving \num{2512} accepted runs across the six complete models. 

For the cloud models, we repeat the same procedure but since we cannot directly measure power, energy usage is derived.  For each query, we measure: wall-clock duration $T_q$, time to first token $\ell_q$, prompt tokens, visible completion tokens, hidden reasoning tokens, cached tokens, attempt count, and finish reason. We estimate energy use under low, central, and high serving-energy scenarios \(s\), reflecting alternative assumptions about request concurrency and accelerator utilization (see SI 2). 

Following \cite{jegham2025hungry}, a request occupying $G$ of a node's $N$ accelerators alongside $B-1$ others draws a share of node power; with $G = N = 8$, 
\begin{equation}
U^{\mathrm{acc}}_s = \frac{G\,D_s}{N\,B_s},\qquad
U^{\mathrm{non}}_s = \frac{D_{\mathrm{non}}}{B_s},\qquad
\Pi_{v,s} = \bigl(P_{\mathrm{acc}}U^{\mathrm{acc}}_s + P_{\mathrm{non}}U^{\mathrm{non}}_s\bigr)\gamma_v ,
\label{eq:power}
\end{equation}
\begin{equation}
E_{q,s} \;=\; \frac{\Pi_{v,s}\,T_q}{3600},
\qquad\text{with}\qquad
e_{q,s} \;=\; \frac{1000\,E_{q,s}}{n_{\mathrm{out}}} ,
\label{eq:cloud}
\end{equation}
where $P_{\mathrm{acc}} = \SI{5600}{W}$ (8 $\times$ H100 SXM), $P_{\mathrm{non}} = \SI{4600}{W}$ (a \SI{10.2}{kW} node less the accelerators), $D_{\mathrm{non}} = 0.5$, and $\gamma_v$ is the power usage effectiveness (PUE) of the provider  (AWS \num{1.14}, Azure \num{1.12}, Google \num{1.09}).  To derive energy per query for a given scenario $E_{q,s}$, we use measured query duration \(T_q\) directly rather than reconstructing it from time to first token and generation rate (as used in \cite{jegham2025hungry}). For our primary analysis, we use the central serving-energy scenario - see SI 2 for derived power usage per model.

The three flagship models (Claude-Opus-4.8 (AWS), GPT-5.5 (Azure),
Gemini-3.1-Pro (GCP)) are queried through a common enterprise gateway using the same 14 prompts and 15 repetitions per prompt. We set \verb|max_tokens| = \num{4096} to match the local-model analysis above, stream responses, disable gateway caching, and impose a \SI{400}{s} stream deadline. However, model-specific restrictions prevented identical inference settings across providers, as GPT-5.5 does not accept $T=0$ and therefore uses its default temperature, while Gemini continues to report reasoning tokens when \verb|reasoning_effort="none"|. Two calls failed and one was classified as a stall. No calls were served from the gateway cache.

Note that the cloud-model runs were collected in two sessions, and between sessions we corrected the measurement code so that retry backoff time was recorded separately from query duration. Because we cannot verify that runs collected before this correction exclude backoff time, we reran the affected model-prompt cells and exclude the 80 original observations from the analysis. These observations were concentrated among Claude runs, affecting 13 of its 14 prompt cells, and it was important to correct this issue in order to properly determine energy use per query for the Claude model.

\subsection*{Electricity and emissions data}
Our data for estimating marginal emission rates are obtained from the Energy Information Administration (EIA) and Environmental Protection Agency (EPA) and span 2022-2025.  EIA-930 reports hourly demand (load) and aggregate generation by technology type for US balancing authorities. Hourly emissions data are obtained from EPA's Continuous Emissions Monitoring System (CEMS) for all power plants greater than 50 MW.  These plant-level emissions data are then aggregated to the corresponding balancing authority based on location. We exclude 23 balancing authorities for which the required hourly data are insufficient to estimate marginal emission rates, primarily because hourly load data are unavailable or because no CEMS-covered generating units can be matched to the balancing authority. Our final sample consists of 43 balancing authorities, where for each balancing authority $i$ and for each hour $t$, we have carbon emissions $E_{it}$ (in short tons), generation by technology $G_{it}$ (in units of MWh), and load $L_{it}$ (in units of MWh). 
Note that because balancing authorities exchange electricity with neighboring regions \cite{fell2021regional}, these estimates capture the relationship between local load and emissions from generators located within the balancing authority's spatial footprint. 
\subsection*{Marginal emissions estimation}
We follow similar work in \cite{Kaff:2013}, \cite{Novan:15}, \cite{CFM:2018},  \cite{bajo2025decarbonization} and many others, who use high-frequency data to estimate marginal emission responses. To estimate marginal carbon emissions per unit of load for different load levels, we regress hourly carbon emissions on hourly load. 
To allow for balancing-authority specific effects of all control variables, we estimate each balancing authority separately. 

For each BA $i$, we separately estimate the following regression specification: 
\begin{equation}\label{spec_cubic}
E_{it} = \beta_{1i} L_{it}+\beta_{2i}L^2_{it}+\beta_{3i}L^3_{it}+\delta_{mh}+\eta_{ym}+\gamma_{yh}+\epsilon_{it}
\end{equation}
\noindent where hourly carbon emissions $E_{it}$ are regressed on a cubic of hourly electricity load $L_{it}$, month-by-hour fixed effects $\delta_{mh}$, year-by-month fixed effects $\eta_{ym}$, and year-by-hour fixed effects $\gamma_{yh}$. Standard errors are clustered at the month-year level, to allow for arbitrary within-month serial correlation.  Month-by-hour fixed effects $\delta_{mh}$ are a key control variable for causal interpretation of the load coefficients, as they control for any systematic, unobservable within-month hourly trends that may be correlated with load and emissions (e.g. seasonal patterns of maintenance). Year-by-month fixed effects $\eta_{ym}$ similarly control for any systematic, unobservable trends evolving over the sample period (e.g. changes in fuel prices or regulation). Finally, we also include year-by-hour fixed effects $\gamma_{yh}$ to allow for hour-specific trends over time, which is particularly important for BA's with growing solar generation.  Thus, within the same month, on the same hour of the day, we are statistically comparing emissions across two otherwise similar hours, but for the difference in hourly electricity demand.
The coefficients of interest, $\beta_{1i}$, $\beta_{2i}$, and $\beta_{3i}$, determine the marginal effect of a 1 MWh increase in load on hourly emissions (measured in short tons) in balancing authority \(i\):

\begin{equation}\label{marg_effect}
\frac{\partial E_{it}}{\partial L_{it}}=\beta_{1i}+2\beta_{2i}L_{it}+3\beta_{3i}L_{it}^2    
\end{equation}
Note that because of the cubic specification, the marginal effect is a function of load $L_{it}$, and over the observed range of load we expect (and find) $\frac{\partial E_{it}}{\partial L_{it}} >0$. 

Next, to estimate marginal emissions and generation responses by hour of day $h$ for Fig. 5, we use an interaction model to regress hourly emissions $E_{it}$ (or generation $G_{it}$) on hourly load $L_{it}$ 
\begin{equation}\label{spec_hourly}
E_{it} = \sum_{h=1}^{24} \beta^h_{i} L_{it}+\delta_{mh}+\eta_{ym}+\gamma_{yh}+\epsilon_{it}
\end{equation}
Here, each $\beta_i^h$ represents the marginal effect of load on emissions in a specific balancing authority for a specific hour (e.g. 9am in MISO).   For each balancing authority in our sample, we estimate 24 hour-specific marginal-effect coefficients. 

\subsection*{Emissions per query and emissions savings calculations}
To determine emissions per query in Fig. 1, we calculate the product of marginal emissions (tons/MWh) in each BA and electricity use per query (MWh/query) in each model, to give us emissions per query in each BA for each model (Fig. 1). To determine the marginal emissions in each BA, we construct a load-weighted marginal CO$_2$ emission rate using the hour-specific marginal emissions estimates above. For each balancing authority \(i\), we weight each estimated hour-specific marginal emission coefficient \(\hat{\beta}_i^h\) by the share of total electricity load occurring during hour \(h\):

\begin{equation}\label{weight}
w_i^h =
\frac{\sum_{t:h(t)=h} L_{it}}
{\sum_t L_{it}},
\qquad h=1,\ldots,24.
\end{equation}
where \(w_i^h\) is the share of total load in balancing authority \(i\) occurring during hour \(h\). This weighting gives greater  importance to marginal emission rates during hours in which a larger share of electricity is consumed. The load-weighted  marginal emission rate is then:

\begin{equation}\label{load_weight}
ME_i^{LW} = \sum_{h=1}^{24}
w_i^h \hat{\beta}_i^h.
\end{equation}

To generate the emissions per query in Fig. 1, we multiply the marginal emission rate for each balancing authority in equation \ref{load_weight} by either the metered energy usage per query for local models in equation \ref{eq:local}, or the derived energy usage per query for cloud models in equation \ref{eq:cloud}.

To determine the potential carbon emissions savings in  Fig.2, Scenario 1 represents the marginal carbon emissions from a query in PJM run at 8pm on the most emissions-intensive models by type (4 billion parameters for local models, GPT-5.5 for cloud models). Scenario 2 answers the query at the least emissions-intensive time of day. Scenario 3 answers the query in the same way as Scenario 2 but with the least-emissions intensive model by type. Scenario 4 answers the query in the same way as Scenario 3 but in BPAT. Finally, for the cloud models, Scenario 5 answers the query in the same way as Scenario 4 but on a local model.

\subsection*{Climate damages relative to private electricity expenditures}

To compare the climate damages associated with AI inference with the private electricity expenditures required to serve the same computation in Figure 6, we construct an external-climate-cost-to-electricity-expenditure ratio for each balancing authority. 

For balancing authority $i$, the climate external cost per query is:
\begin{equation}
EC_i = e_q \times ME_i^{LW} \times \kappa \times SCC,
\end{equation}
where $e_q$ denotes electricity consumption per query in MWh, $ME_i^{LW}$ is the load-weighted average marginal CO$_2$ emission rate defined above, measured in tons of CO$_2$ per MWh, $\kappa=0.907185$ converts U.S. tons to metric tons, and $SCC$ is the social cost of carbon in dollars per metric ton of CO$_2$. We use an SCC of \$185 per metric ton of CO$_2$ in 2020 U.S. dollars, equivalent to approximately \$224 in 2024 dollars after adjusting for inflation.

Private electricity expenditure per query is:
\begin{equation}
PC_i = 1000 \times e_q \times P_i,
\end{equation}
where $P_i$ is the 2024 average retail electricity price in balancing authority $i$, measured in dollars per kWh and calculated from utility revenues and sales reported in EIA-861. Our primary specification uses the average retail price across all customer classes.

We define the external-cost-to-electricity-expenditure ratio as:
\begin{equation}
R_i = \frac{EC_i}{PC_i} = \frac{ME_i^{LW} \times \kappa \times SCC}{1000P_i}.
\end{equation}

A value of $R_i=1$ indicates that estimated climate damages equal the private electricity expenditure associated with serving the query, while values above one indicate that climate damages exceed private electricity expenditures. Because the same query-level electricity requirement enters both the numerator and denominator, $e_q$ (emissions per query) cancels from the ratio. The ratio therefore hinges on the marginal emissions intensity of electricity generation, the social cost of carbon, and local electricity prices.

\bibliography{bibliography}

@article{jegham2025hungry,
  title={How hungry is ai? benchmarking energy, water, and carbon footprint of llm inference},
  author={Jegham, Nidhal and Abdelatti, Marwan and Koh, Chan Young and Elmoubarki, Lassad and Hendawi, Abdeltawab},
  journal={arXiv preprint arXiv:2505.09598},
  year={2025}
}

@article{CFM:2018,
  title={Location, location, location: The variable value of renewable energy and demand-side efficiency resources},
  author={Callaway, Duncan S and Fowlie, Meredith and McCormick, Gavin},
  journal={Journal of the Association of Environmental and Resource Economists},
  volume={5},
  number={1},
  pages={39--75},
  year={2018},
  publisher={University of Chicago Press Chicago, IL}
}

@ARTICLE{Novan:15,
  AUTHOR =       {Kevin M. Novan},
  TITLE =        {Valuing the Wind: Renewable Energy Policies and Air Pollution Avoided},
  JOURNAL =      {American Economic Journal: Economic Policy},
  YEAR =         {2015},
   volume =       {7},
   number = {3},
  pages =        {291-326},
}

@article{Kaff:2013,
  title={Emissions Savings from Wind Power Generation in {Texas}},
  author={Kaffine, Daniel T and McBee, Brannin J and Lieskovsky, Jozef},
  journal={Energy Journal},
  volume={34},
  number={1},
  pages={155--175},
  year={2013}
}

@article{fell2021regional,
  title={Regional disparities in emissions reduction and net trade from renewables},
  author={Fell, Harrison and Johnson, Jeremiah X},
  journal={Nature Sustainability},
  volume={4},
  number={4},
  pages={358--365},
  year={2021},
  publisher={Nature Publishing Group UK London}
}

@article{holland2022marginal,
  title={{Why marginal CO2 emissions are not decreasing for US electricity: Estimates and implications for climate policy}},
  author={Holland, Stephen P and Kotchen, Matthew J and Mansur, Erin T and Yates, Andrew J},
  journal={Proceedings of the National Academy of Sciences},
  volume={119},
  number={8},
  pages={e2116632119},
  year={2022},
  publisher={National Academy of Sciences}
}

@techreport{muller2026measuring,
  title={Measuring the impact of data centers in the United States economy: Monetary damage from air pollution and greenhouse gas emissions},
  author={Muller, Nicholas Z},
  year={2026},
  institution={National Bureau of Economic Research}
}

@article{bajo2025decarbonization,
  title={Decarbonization and electricity price vulnerability},
  author={Bajo-Buenestado, Raul and Bento, Antonio M and Kaffine, Daniel and Marmarelis, Zissis E},
  journal={Nature Sustainability},
  volume={8},
  number={2},
  pages={170--181},
  year={2025},
  publisher={Nature Publishing Group UK London}
}

@article{rennert2022comprehensive,
  title={Comprehensive evidence implies a higher social cost of CO2},
  author={Rennert, Kevin and Errickson, Frank and Prest, Brian C and Rennels, Lisa and Newell, Richard G and Pizer, William and Kingdon, Cora and Wingenroth, Jordan and Cooke, Roger and Parthum, Bryan and others},
  journal={Nature},
  volume={610},
  number={7933},
  pages={687--692},
  year={2022},
  publisher={Nature Publishing Group UK London}
}

@misc{openai2025economicanalysis,
  author = {{OpenAI}},
  title  = {OpenAI's New Economic Analysis},
  year   = {2025},
  month  = jul,
  url    = {https://openai.com/global-affairs/new-economic-analysis/},
  note   = {Accessed September 21, 2026}
}

@techreport{epa2020egrid,
  author      = {{U.S. Environmental Protection Agency}},
  title       = {The Emissions \& Generation Resource Integrated Database: Technical Support Document for eGRID with Year 2018 Data},
  institution = {U.S. Environmental Protection Agency, Office of Atmospheric Programs, Clean Air Markets Division},
  year        = {2020},
  month       = jan,
  url         = {https://www.epa.gov/sites/default/files/2020-01/documents/egrid2018_technical_support_document.pdf}
}

@article{guidi2026assessing,
  title={Assessing the Carbon Emissions and Energy Consumption of US Hyperscale Data Centers},
  author={Guidi, Gianluca and Dominici, Francesca and Squartini, Tiziano and Sprinkle, Callaway and Gilmour, Jonathan and Butler, Kevin and Bell, Eric and Delaney, Scott and Bargagli-Stoffi, Falco J},
  journal={Nature Sustainability},
  year={2026}
}

@article{you2026quantifying,
title = {Quantifying the regional carbon impact of AI inference workloads: A grid-aware NeuroPhysical network approach},
journal = {Applied Energy},
volume = {422},
pages = {128298},
year = {2026},
issn = {0306-2619},
doi = {https://doi.org/10.1016/j.apenergy.2026.128298},
url = {https://www.sciencedirect.com/science/article/pii/S0306261926009529},
author = {Haiyang You and Chengwei Lou and Yue Zhou and David Flynn and Jin Yang}
}

@misc{bistline2026aiemissions,
  author = {John Bistline},
  title  = {Watt's the Difference? Your Chatbot's Carbon Depends on Its ZIP Code},
  year   = {2026},
  month  = sep,
  url    = {https://watershed.com/blog/ai-emissions-by-region},
  note   = {Watershed, accessed September 23, 2026}
}

@misc{aubakirova2025stateofai,
  author = {Malika Aubakirova and Alex Atallah and Chris Clark and Justin Summerville and Anjney Midha},
  title  = {State of AI: An Empirical 100 Trillion Token Study with OpenRouter},
  year   = {2025},
  month  = dec,
  url    = {https://openrouter.ai/state-of-ai},
  note   = {OpenRouter and Andreessen Horowitz}
}

@misc{eia2025electricityconsumption,
  author = {{U.S. Energy Information Administration}},
  title  = {After More Than a Decade of Little Change, U.S. Electricity Consumption Is Rising Again},
  year   = {2025},
  month  = may,
  url    = {https://www.eia.gov/todayinenergy/detail.php?id=65264},
  note   = {Today in Energy, May 13, 2025}
}

@article{holland2016there,
  title={Are there environmental benefits from driving electric vehicles? The importance of local factors},
  author={Holland, Stephen P and Mansur, Erin T and Muller, Nicholas Z and Yates, Andrew J},
  journal={American Economic Review},
  volume={106},
  number={12},
  pages={3700--3729},
  year={2016},
  publisher={American Economic Association 2014 Broadway, Suite 305, Nashville, TN 37203}
}

@article{fell2026spatio,
  title={Spatio-Temporal Heterogeneity in the Environmental Value of Renewable Energy},
  author={Fell, Harrison and Kaffine, Daniel},
  journal={The Energy Journal},
  pages={01956574261462659},
  year={2026},
  publisher={SAGE Publications Sage CA: Los Angeles, CA}
}

@article{nguyen2024towards,
  title={Towards sustainable large language model serving},
  author={Nguyen, Sophia and Zhou, Beihao and Ding, Yi and Liu, Sihang},
  journal={ACM SIGENERGY Energy Informatics Review},
  volume={4},
  number={5},
  pages={134--140},
  year={2024},
  publisher={ACM New York, NY, USA}
}

@inproceedings{dodge2022measuring,
  title={Measuring the carbon intensity of AI in cloud instances},
  author={Dodge, Jesse and Prewitt, Taylor and Tachet des Combes, Remi and Odmark, Erika and Schwartz, Roy and Strubell, Emma and Luccioni, Alexandra Sasha and Smith, Noah A and DeCario, Nicole and Buchanan, Will},
  booktitle={Proceedings of the 2022 ACM conference on fairness, accountability, and transparency},
  pages={1877--1894},
  year={2022}
}

@article{gagnon2022short,
  title={Short-run marginal emission rates omit important impacts of electric-sector interventions},
  author={Gagnon, Pieter J and Bistline, John ET and Alexander, Marcus H and Cole, Wesley J},
  journal={Proceedings of the National Academy of Sciences},
  volume={119},
  number={49},
  pages={e2211624119},
  year={2022},
  publisher={National Academy of Sciences}
}

@techreport{rhodium_2026,
author={King, Ben and Bower, Galen and Case, Parker and Kolus, Hannah and Larsen, John and Pastorek, Nathan and Wiens, Nathan},
title={Estimating the US Energy System
Impacts of the Bipartisan American
Affordability and Jobs Act},
institution={Rhodium Group},
year={2026},
}

\end{document}